\documentclass[runningheads]{llncs}
\usepackage{amsmath}
\usepackage{graphicx}
\usepackage{amssymb}
\usepackage{multirow}
\usepackage{booktabs} 
\usepackage{url}
\usepackage[ruled,vlined]{algorithm2e} 

\SetKwInput{KwInput}{Input}                
\SetKwInput{KwOutput}{Output}              
\SetKwFor{For}{for}{do}{endfor for}
\SetKwIF{If}{ElseIf}{Else}{if}{then}{else if}{else}{endif}
\begin{document}

\title{CoG: a Two-View Co-training Framework for Defending Adversarial Attacks on Graph}

\titlerunning{CoG: a Two-View Co-training Framework on Graph}
\author{Xugang Wu\inst{1}\and
Huijun Wu\inst{1}\and
Xu Zhou\inst{1}\and
Kai Lu\inst{1}}
\authorrunning{Wu et al.}
%
\institute{National University of Defense Technology\\
\email{xugangwu95@gmail.com}}

\maketitle              

\begin{abstract}

Graph neural networks (GNNs) exhibit remarkable performance in graph data analysis. However, the robustness of GNN models remains a challenge. As a result, they are not reliable enough to be deployed in critical applications. Recent studies demonstrate that GNNs could be easily fooled with adversarial perturbations, especially structural perturbations. Such vulnerability is attributed to the model’s excessive dependence on the structure information to make predictions. To achieve better robustness, it is desirable to build the prediction of GNNs with more comprehensive features. Graph data, in most cases, has two views of information, namely structure information and feature information. In this paper, we propose CoG, a simple yet effective co-training framework to combine these two views for the purpose of robustness. CoG trains sub-models from the feature view and the structure view independently and allows them to distill knowledge from each other by adding their most confident unlabeled data into the training set. The orthogonality of these two views diversifies the sub-models, thus enhancing the robustness of their ensemble. We evaluate our framework on three popular datasets, and results show that CoG significantly improves the robustness of graph models against adversarial attacks without sacrificing their performance on clean data. We also show that CoG still achieves good robustness when both node features and graph structures are perturbed.

\end{abstract}

\section{Introduction}
Graph Neural Networks~(GNNs) have achieved remarkable success in the learning representations for graph-structured data across many domains, such as citation networks, biological networks, and social networks. Among these models, Graph Convolutional Network (GCN) and its variants~\cite{Kipf2017,Velickovic2018,Hamilton2017,Fey2019} have gained much attention because of their performance and efficiency. However, recent studies demonstrate that these message-passing-based models suffer from adversarial perturbations. The accuracy of the learned model drops drastically when the attacker conducts unnoticeable modifications on the graph data~\cite{dai2018adversarial,zugner2018adversarial,Bojchevski2019,zugner2019adversarial}. Further results~\cite{Wu2019,zugner2019adversarial} illustrate that structural perturbations lead to more effective attacks compared to feature perturbations. 

\begin{figure}
\centering
\includegraphics[width=0.6\textwidth]{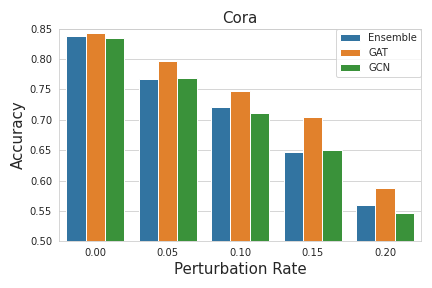}
\caption{Performance of ensemble of GCN and GAT under Metattack.} \label{fig1}
\vspace{-10pt}
\end{figure}

Some defense techniques are proposed to enhance the robustness of GNNs against structural perturbations. The general principle is to eliminate the negative impacts brought by perturbed edges based on certain prior knowledge. For example, some defense methods~\cite{Zhang2020,chen2021enhancing,Luo2020,Wu2019} assume that all the connected nodes should be similar in the feature space. Based on it, they remove or pay less attention to edges that connect dissimilar nodes.  Other methods~\cite{Jin2020,Entezari2020} observe that structural perturbations tend to affect the high-rank portion of the graph. Therefore, they replace the graph topology with its low-rank approximation to purify the graph. However, these heuristic knowledge may not always hold in all scenarios. For example, in heterophilic graphs, removing edges between dissimilar nodes could impair the model's performance. Another drawback of heuristic defenses is their weakness against adaptive attacks. Among these methods, few take adaptive attacks into consideration. In this paper, instead of introducing heuristic knowledge,  we propose to defend against poisoning attack by learning an ensemble model that fully exploits the feature and structure information. 

Ensemble method is a prevalent way to defend against adversarial attacks in the field of computer vision. Recent works~\cite{Pang2019,Kariyappa2019,Yang2020} point out that output diversity is the key to the success of these ensemble-based defense methods.  However, applying ensemble techniques to GNN straightforwardly may not improve the robustness of the ensemble model. Fig.~\ref{fig1} shows how the ensemble of GCN and GAT performs under Metattack. Metattack can transfer between GCN and GAT well, leading to poor robustness of their ensemble. 

In this paper, We argue that introducing diversified models that share different views of graph data will help to improve the overall robustness of the ensemble. Compered to image data, graph data have two orthogonal views: the node feature view and the graph structure view. The orthogonality of these two views makes it difficult for adversarial attacks to transfer between them. To fully exploit this property, our work uses a two-view co-training framework~(CoG) to learn an ensemble of sub-models from both feature view and structure view. Co-training~\cite{blum1998combining,wang2007analyzing} is a simple yet effective technique for learning an ensemble of sub-models in two different views under semi-supervised setting. It trains a separate classifier for each view and adds the most confident predictions of each sub-model to the training data set. During this process, sub-models distill knowledge from each other and get strengthened. Nevertheless, applying vanilla co-training framework to graph data meets two challenges: (1) Co-training uses the softmax outputs as the indicator of sub-models’ confidence. However, this can be inaccurate since  neural networks are often miscalibrated, especially when sub-models are heterogenous. (2) Co-training selects unlabeled data simply based on their confidence. If dominant classes exist, the co-training process will amplify the imbalance of classes and force the sub-models to overfit to the dominant class. To address these issues, we use temperature scaling to calibrate the outputs, and enforce the constant of class distribution when adding predictions during the co-training process.

Experiment results show that the diverse outputs between feature and structure view prevent adversarial examples from transferring between them, thus making their ensembles more robust against adversarial attacks, even in the scenario of adaptive attacks. To conclude, our contributions are summarized as follows:
\begin{itemize}
    \item To employ the orthogonality between feature and structure views, we propose a two-view co-training framework~(CoG) to combine the feature information and the structure information during the training process.
    \item We show that model calibration and class balancing play significant roles in enabling effective co-training.
    \item Experiments show that CoG can achieve better robustness against poisoning attacks compared to other state-of-the-art methods, without sacrificing the performance on clean data.
    \item We conduct adaptive attacks and results indicate that our method can still work well under the adaptive setting.
\end{itemize} 

\section{Related work}
\subsection{Attack and defense on graph data}
Despite the great success of GNNs, recent works indicate that these graph-based models are vulnerable to unnoticeable modifications~\cite{zugner2018adversarial,zugner2019adversarial,dai2018adversarial}. Nettack~\cite{zugner2018adversarial} conducts its attack on a surrogate model and ensures their modifications to be unnoticeable via considering the degree distribution and feature co-occurrence. RL-S2V~\cite{dai2018adversarial} applies reinforcement learning to generate adversarial examples. Metattack~\cite{zugner2019adversarial} addresses the global attack problem. It uses the meta-gradient to generate a perturbed graph that leads to an overall decrease in model's performance. All these attack methods suggest that structural attacks are more effective than feature attacks when applied to graph data. 

Several techniques have been proposed to defend these topological attacks~\cite{Wu2019,Zhang2020,Jin2020,Jin2020_2,chen2021enhancing,Luo2020}. GCN-Jaccard~\cite{Wu2019} assumes that connected nodes should have similar features so that edges between dissimilar nodes are more likely to be the perturbed edges. Based on this assumption, they calculate the Jaccard similarity scores between all connected nodes and remove edges that connect nodes with scores below the threshold. Similarly, GNNGuard~\cite{Zhang2020} employs the attention mechanism to assign higher weights to the edges between similar nodes and less to the edges between unrelated nodes. Other works stem from the observation that structural attacks tend to affect the high-rank portion of a graph. GCN-SVD~\cite{Entezari2020} is proposed to purify the perturbed graph via replacing it with its low-rank approximation, while Pro-GNN~\cite{Jin2020} introduces a regularization term to generate a low-rank and sparse graph during the training process.  However, these methods rely on the validity of their heuristic knowledge. SimP-GCN~\cite{Jin2020_2} attempts to integrate the structure information and node features by combining the kNN graph and the original graph. Nevertheless, the scoring function that balances these two graphs merely depends on the hidden representation so that the graph used for learning could be unstable, thus leading to high variance for the results.

The most related defense method is UM-GNN~\cite{shanthamallu2021uncertainty}. UM-GNN is proposed to learn a feature-based model via distilling knowledge from GNN models using an uncertainty matching strategy. However, its knowledge distillation is one-directional: only the feature-based model can distill knowledge from GNN models. Consequently, GNN models cannot enhance itself using the information from the feature-based model. As the perturbation rate grows, the knowledge transferred from GNN models gets less effective, which impairs the performance of UM-GNN.

\subsection{Ensemble training for enhanced robustness}\label{ensemble}
Although ensemble training was initially proposed to improve models' performance~\cite{Hansen1990,Breiman1996,dietterich2000ensemble,Kuncheva2003}, a recent line of works~\cite{Pang2019,Kariyappa2019,Yang2020} show that it can be applied to enhance adversarial robustness. The intuition behind these methods is that a small overlap between adversarial subspace (Adv-SS) of different sub-models can prevent adversarial attacks from transferring between sub-models. Pang et al.~\cite{Pang2019} employ an adaptive diversity-promoting regularizer to encourage diversity among non-maximal predictions. Kariyappa et al.~\cite{Kariyappa2019} propose diversity training to reduce the correlation of loss functions between sub-models. Yang et al.~\cite{Yang2020} distill the non-robust features from each sub-model and teach other sub-models to be robust against these non-robust features. Although these defense approaches have been well studied in image recognition tasks, their application in graph-based tasks remains to be explored. 
\vspace{-5pt}


\section{Preliminary study} \label{sec3}

In this section, we present why two-view co-training training is a desirable defense technique for graph data. As is discussed in the previous section, output diversity plays a crucial role in the robustness of the ensemble model. Recent work~\cite{tramer2017space} shows that adversarial examples are more likely to transfer between models that have a large adversarial subspace~(Adv-SS) overlap. Sub-models with diverse outputs have a smaller Adv-SS overlap, making it harder for attackers to craft adversarial examples that fool both two models~\cite{Pang2019,Kariyappa2019,Yang2020}. 

To minimize the Adv-SS overlap, using different aspects of the input data is a promising approach.  Graph data contains two orthogonal views: a node feature view and a graph structure view. Models in these two views naturally share little adversarial subspace. An attack on node features would barely affect a structure-based model and vice versa, which implies the potential for improving robustness via an ensemble of sub-models from these two views. Furthermore, from attackers' perspective, conducting attacks on the feature view is more difficult due to the following reasons: (1) in most graph data, node features are sparse and high-dimensional, which makes the perturbations on node features detectable; (2) in most cases, node features are discrete and practically interpretable, which causes restrictions on feature modifications. Suppose an attacker attempts to attack a social network via modifying a person's birthday, his/her age should be simultaneously changed to ensure the data validity. Such restrictions are difficult to model in a differentiable manner, which constitutes a challenge for gradient-based attacks.

Although existing message-passing-based GNN models are designed to integrate the feature and structure information of graph data, they do not pay enough attention to the feature information. In an empirical study, Jin et al.~\cite{Jin2020} point out that GCN prefers to preserve structure information rather than feature information during the message passing process. We call it a structure-dominant model. For these models, the node's representation is highly related to its neighboring structure rather than its feature. This behavior is consistent with the empirical conclusions in previous work ~\cite{Wu2019}: (1) Topological attacks are more effective than feature attacks when attacking models like GCN; (2) Attackers tend to connect nodes with dissimilar node features to achieve a successful attack. 

Therefore, to fully exploit the feature information and achieve better robustness, we propose a two-view co-training framework, named Co-training on Graph~(CoG), to learn a robust ensemble of the feature-dominant models and the structure-dominant models.

\begin{figure*}
\centering
\includegraphics[width=0.8\textwidth]{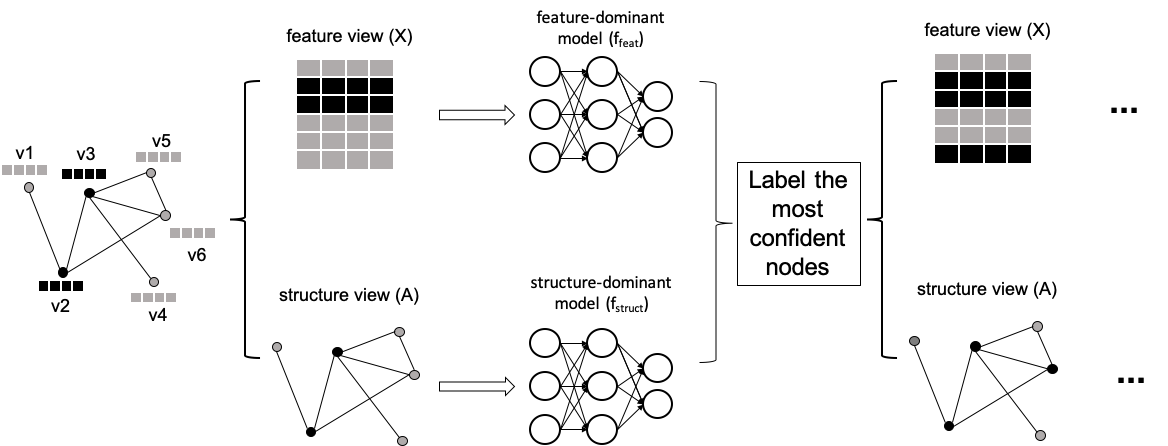}
\caption{The overall framework of CoG. Graph data is seperated into a feature view and a structure view. During the co-training process, sub-models from each view label their most confident nodes and add them into the training set.} \label{fig2}
\end{figure*}

\section{Proposed approach}
\subsection{Problem formulation}
In this paper, we focus on the semi-supervised node classification problem. Let $\mathcal{G}=(\mathcal{V}, \mathcal{E}, X)$ be a graph with n nodes, where $\mathcal{V}$ is the set of nodes $\left\{v_{1}, \cdots, v_{n}\right\}$ with $|\mathcal{V}|=n, \mathcal{E}$ is the set of edges, and $X= \left[x_{1}, x_{2}, \cdots, x_{n}\right]^{T} \in \mathbb{R}^{n \times m}$ is a feature matrix. Then we can separate $\mathcal{G}$ into two views.
In the node feature view, for each node $v \in \mathcal{V}$, its feature $x_{v} \in \mathbb{R}^{m}$ is a $m$-dimensional row vector. In the graph structure view,  the adjacency matrix $A\in \mathbb{R}^{n \times n}$ can be formulated by setting $A_{i j}=1$ if $\left(v_{i}, v_{j}\right) \in \mathcal{E}$ and $A_{i j}=0$, otherwise. In the semi-supervised node classification problem, nodes are separated into two sets $\mathcal{V} = \mathcal{S} \cup \mathcal{U}$, where nodes in $\mathcal{S}$ are labeled and nodes in $\mathcal{U}$ are not. Our goal is to learn an high-performance ensemble of a feature-dominant model $f_{feat}$ and a structure-dominant model $f_{struct}$, using both the labeled and unlabeled data from $\mathcal{V}$.

\subsection{The overall framework} 
Graph data can be separated into two views, i.e., the feature view and the structure view. These two views provide different information about the nodes. The feature view describes nodes' intrinsic properties, while the structure view tells the relationship between each node pair. Furthermore, they meet the requirements\cite{Kuncheva2003} of a co-training framework well. (1) Models trained in any of these two views can classify the nodes effectively. (2) Information provided by these two views is independent so that knowledge distilled from one model can promote the performance of the other model. Therefore, we apply the two-view co-training framework on graph data~(CoG) and learn an ensemble of sub-models in these two views.

The overall framework of CoG is shown in Fig.~\ref{fig2}. For each node $v$, we separate its information into a feature view $X$ and a structure view $A$. After that, two classifiers are trained separately for each view. Feature-dominant models primarily use node features as its input to classify the nodes, while structure-dominant models usually use the graph structure as its clues for node labels. In each iteration, the most confident unlabeled nodes from each view are added to the training set. After a preset number of iterations, we obtain two well-trained models, $f_{feat}$ and $f_{struct}$, and generate an ensemble of them by averaging their predictions. In the remaining part of this section, we introduce more details about the framework, including (1) the structure-dominant models and feature-dominant models we use to learn representation for nodes; (2) the techniques we use to improve the performance of the co-training framework, including model calibration and class balancing.

\subsection{Structure-dominant models}
\subsubsection{GCN} Graph convolutional network~(GCN)~\cite{Kipf2017} achieves remarkable success on learning representation for graph data. Its message-passing mechanism learns the representation of each node via propagating information and aggregating with a permutation invariant function. Given a graph $\mathcal{G}=(X, A)$, the updates of node embeddings can be derived by the following formulation:
\begin{equation}
    X^{(l+1)}=\sigma\left(\hat{A} X^{(l)} W^{(l)}\right),
\end{equation}
where $\hat{A}$ is the normalized adjacency matrix with self loops $\hat{A} = \tilde{D}^{-\frac{1}{2}}(A+I)\tilde{D}^{-\frac{1}{2}}$, $I$ is the identity matrix, $\tilde{D}_{i i}=\sum_{j} \tilde{A}_{i j}$, $W^{(l)}$ is the feature transformation matrix of the $l$th layer, and $\sigma(\cdot)$ is the activation function, e.g., ReLU(x). Although GCN takes both features and graph structure as its input, the representation it learns for each node largely depends on the local graph structure of the node.

\subsubsection{Spectral-based method (S-MLP)}
Spectral-based method is another effective method to distill structure information for each node. Utilizing the spectral decomposition of the graph laplacian to represent both local and global structure has been thoroughly studied in the graph spectral theory~\cite{chung1997spectral,von2007tutorial}. Given the graph topology $A$, we first compute its eigenvalues and eigenvectors of laplacian by solving:
\begin{equation}
    D^{-1}L \mathbf{y} = \lambda \mathbf{y}
\end{equation}
where $D^{-1}L = I - D^{-1}A$ is the normalized laplacian matrix. Let $\lambda_0, \lambda_1, \cdots, \lambda_k$ be the k smallest eigenvalues of $D^{-1}L$ and $\mathbf{y}_0, \mathbf{y}_1, \cdots, \mathbf{y}_k$ be the respective eigenvectors, each node $u$ can be embedded into a k-dimension space as $a_{u} = (\mathbf{y}_{1u}, \mathbf{y}_{2u}, \cdots, \mathbf{y}_{ku})$. 

To enrich the information of $a_u$, we also compute the k-dimension laplacian eigenmaps of $A^2$. By concatenating them together, we get the enhanced $a_u \in \mathbb{R}^{2 \times k}$. After that, $a_u$ is sent as inputs to an MLP model for classification.

\subsection{Feature-dominant models}
\subsubsection{MLP (F-MLP)} Multi-layer perception~(MLP) model is the simplest while effective method when we consider the node features alone. The layerwise forwarding process of  MLP can be formulated as,
\begin{equation}
    x_{v}^{(l+1)}=\sigma (x_{v}^{l}\Theta^{(l)} + b_l)
\end{equation}
where  $x_{v} \in \mathbb{R}^{m}$ is a $m$-dimensional row vector denotes the features of node $v$, $\Theta^{(l)}$ and $b_l$ are the layer-wise parameters. $\sigma(\cdot)$ denotes the activation function. 

\subsubsection{kNN-based GCN (kNN-GCN)} In kNN-based model, we first construct a feature graph from feature matrix $X$ by conducting k-nearest-neighbor algorithm based on the cosine similarity. For each node pair $(v_i,v_j)$, we calculate their feature similarity as:
\begin{equation}
   \mathrm{s}_{i j}=\frac{\mathrm{x}_{i}^{\top} \mathrm{x}_{j}}{\left\|\mathrm{x}_{i}\right\|\left\|\mathrm{x}_{j}\right\|}.
\end{equation}
After that, the kNN graph $A_{k} = kNN(X)$ can be constructed by connecting the top-k similar node pairs. Finally, the whole graph data ${\mathcal{G} = (X, A_k)}$ is sent to a GCN model to classify the nodes. The structure information here is generated from the node features, so the original graph structure is not used in this model. 

\subsection{Model calibration}
Models' confidence is the most important indicator during the co-training process. On one hand, the co-training framework picks up the most confident predictions in each sub-model. When sub-models attempt to add the same node to the training set, the co-training framework decides which pseudo label to use based on their confidence. On the other hand, in the inference stage,  CoG average each sub-model’s confidence to obtain the predictions of their ensemble. 

Generally, the softmax output of each sub-model is used to measure the confidence. However, \cite{guo2017calibration,teixeira2019graph} show that modern neural networks, including GNN, can be miscalibrated. Since the sub-models are heterogeneous in CoG, this miscalibration can impair the performance of the co-training method. To mitigate the impacts of model's miscalibration, we use the temperature scaling method~\cite{guo2017calibration} to calibrate the output of each sub-model. Given the logit vector $z$, the calibrated prediction is obtained as:
\begin{equation}
   q = \operatorname{softmax}(z/T),
\end{equation}
where the temperature $T$ is learned by optimization with respect to negative log likelihood on the validation set.

\subsection{Class balancing}
In CoG, the most confident unlabelled nodes from each view are added to the training set during the co-training process. The confidence of each node is measured by the softmax outputs of the sub-models. However, this strategy can lead to class imbalance if sub-models perform better on one particular class. What's worse, an imbalanced class distribution will cause the overfitting problem as the co-traning process moves on and finally impair the sub-models' performance. 

To keep a balanced training dataset, the added data should follow the class distribution of the initial training dataset. Formally, suppose we have $N$ inputs with an initial distribution $(N_1, N_2, \cdots, N_C)$, the number of added proposals for each class $C$ is:
\begin{equation}
   N^{add}_{c} =\frac{N_c}{N} \cdot N^{add} 
\end{equation}
where $N^{add}$ denotes the number of data we add to the training data in each iteration. 

The co-training process will stop when reaching the preset number of iterations or there is no more data in the test set. In the inference phase, we average sub-models' output to get the predictions of their ensemble. Compared to other defense methods, the advantages of CoG can be summarized as follows:

\begin{itemize}
    \item Rather than rely on heuristic knowledge to purify the perturbed graph, CoG enhances the robustness of GNNs via knowledge distillation. Therefore, it could be applied to more scenarios given the fact that it does not make any assumptions about the graph.

    \item Compared to existing ensemble approaches, where the knowledge of the graph only flows from GNN to MLP in an one-direction manner\cite{shanthamallu2021uncertainty} or model diversity is not considered~\cite{ensemble}, CoG enables a bi-directional knowledge distillation between GNN and MLP as well as takes the complementation of different types of model into consideration. This enhance the robustness of CoG when the graph structure is heavily perturbed. Furthermore, the distillation process of CoG is dynamic and non-differentiable, making it more difficult for attackers to conduct adaptive attacks.
    
\end{itemize}

\section{Experimental Results}
In this section, we evaluate the performance of CoG on clean data and its robustness against adversarial attacks. In particular, we focus on answering the following questions: 

Q1. ~How does CoG perform on clean data? 

Q2. How does CoG perform under adversarial attacks compared to other state-of-the-art defense methods? 

Q3. ~How do model calibration, the class balancing technique and the hyper-parameters affect  CoG's performance? 

Q4. ~How does CoG perform against adaptive attacks?

\begin{table} 
\centering 
\caption{Statistics of the datasets.}
\resizebox{0.6\linewidth}{!}{%
\begin{tabular}{c c c c c} 
\toprule 
\midrule
& \# of nodes & \# of edges & Classes & Features \\ 
\midrule 
Cora & 2,485 & 5,069 & 7 & 1,433 \\
Citeseer & 2,110 & 3,668 & 6 & 3,703 \\
Pubmed & 19,717 & 44,338 & 3 & 500 \\
\midrule 
\bottomrule 
\end{tabular}}%
\vspace{-10pt}
\label{dataset} 
\end{table}

\subsection{Experiment Setup}
\subsubsection{Datasets} To obtain comparable results, we use three popular citation graphs, i.e., Cora~\cite{mccallum2000automating}, Citeseer~\cite{sen2008collective} and Pubmed~\cite{sen2008collective} to evaluate our model. The basic information of these three graphs is shown in Table~\ref{dataset}. In these three datasets, each node represents a document and edges are the citations between documents. In terms of data splits, we randomly pick 10$\%$ of nodes for training, 10$\%$ for validation and 80$\%$ for testing, following~\cite{Jin2020,zugner2018adversarial,zugner2019adversarial}.

\subsubsection{Baselines}
We compare our model with the following baselines:
\begin{itemize}
    \item GCN~\cite{Kipf2017}: As we describe in the previous sections, GCN is a representative graph neural network proposed by Kipf and Welling.
    \item GCN-SVD~\cite{Entezari2020}: GCN-SVD is a defense method based on low-rank approximation. 
    \item GCN-Jaccard~\cite{Wu2019}: GCN-Jaccard is also a preprocessing-based defense, which removes the edges between most dissimilar nodes to purify the graph.
    \item SimP-GCN~\cite{Jin2020_2}: SimP-GCN adaptively combines the original graph and the kNN graph to capture the similarity between nodes.
    \item Pro-GNN~\cite{Jin2020}: Pro-GNN is a defense method that exploits three properties of real-world graphs: low-rank, sparsity, and feature smoothness.
    \item UM-GNN~\cite{shanthamallu2021uncertainty}: UM-GNN trains a feature-based model based on knowledge transferred from GNN models.
\end{itemize}
    
\subsubsection{Parameter Settings} We implement CoG under the framework of DeepRobust~\cite{li2020deeprobust}, which is a well-known adversarial learning framework for GNNs. Similar to~\cite{Jin2020,chen2021enhancing}, we run each experiment 10 times and report the average performance. For GCN, we use the settings of the original GCN~\cite{Kipf2017}, i.e., a two-layer structure with 16 hidden units. We use a two-layer structure with 32 hidden units for the MLP model. The number of nearest neighbors $k$ we set in kNN-GCN is 50. As for S-MLP, we use the eigenvectors corresponding to the lowest 50 eigenvalues to distill structure information. In the co-training process, each model adds 250 most confident unlabeled data with their pseudo-labels in one iteration. For the baseline models, we take the same experiment settings as in \cite{Jin2020}. We set the learning rate for all sub-models to 0.01, the weight decay to 5e-4, the dropout rate to 0.5, and train for 200 epochs.

As for adversarial attacks, we use Metattack~\cite{zugner2019adversarial} which is an effective poisoning attack method on graphs. It treats the graph as a hyperparameter and modifies the graph to increase learning loss via meta-gradient. The edge perturbation rate is set to $\{5\%,10\%,15\%,20\%\}$. We use the same random seed as~\cite{Jin2020} to make fair comparisons with their reported results. When comparing with UM-GNN, we also evaluate CoG under PGD attack~\cite{xu2019topology}. The results on PGD attack are in the supplementary material.

\begin{table}[t] 
\centering 
\caption{Node classification accuracy ($\%$) on clean graphs and perturbed graphs.}
\resizebox{0.8\textwidth}{!}{%
\begin{tabular}{c c c c c c c} 
\toprule 
\midrule
\multirow{2}{*}{\textbf{Datasets}}& \multirow{2}{*}{\textbf{Model}} &\multicolumn{5}{c}{\textbf{Perturb Rate$(\%)$}} \\ 
\cmidrule(l){3-7} 
& & 0 & 5 & 10 & 15 & 20 \\ 
\midrule 
\multirow{8}{*}{Cora} 
& GCN & \textbf{83.50±0.44} & 76.55±0.79 & 70.39±1.28 & 65.10±0.71 & 59.56±2.72\\ 
& GCN-SVD & 80.63±0.45 & 78.39±0.54 & 71.47±0.83 & 66.69±1.18 & 58.94±1.13\\ 
& GCN-Jaccard & 82.05±0.51 & 79.13±0.59 & 75.16±0.76 & 71.03±0.64 & 65.71±0.89\\ 
& SimP-GCN & 81.81±0.62 & 76.43±1.98 & 73.27±1.93 & 70.75±3.98 & 66.63±6.87\\ 
& Pro-GNN & 82.98±0.23 & \textbf{82.27±0.45} & \textbf{79.03±0.59} & \textbf{76.40±1.27} & \textbf{73.32±1.56}\\ 
\cmidrule(l){2-7}
& GCN+F-MLP & 84.09±0.59 & \textbf{83.48±0.43} & \textbf{82.88±0.83} & \textbf{81.01±0.57} & 76.70±0.63\\ 
& GCN+kNN-GCN & \textbf{84.27±0.31} & 83.16±0.28 & 82.54±0.29 & 80.57±0.40 & \textbf{76.86±0.75}\\ 
\cmidrule(l){2-7}
& S-MLP & 78.30±0.17 & 76.25±0.23 & 72.95±0.36 & 67.31±0.40 & 54.56±0.25\\
& S-MLP+F-MLP & \textbf{83.59±0.30} & \textbf{84.25±0.33} & \textbf{83.52±0.31} & \textbf{82.95±0.59} & \textbf{80.75±0.59}\\ 
& S-MLP+kNN-GCN & 
82.80±0.33 & 83.85±0.27 & 83.15±0.58 & 82.74±0.23 & 80.74±0.31\\ 
\midrule 
\midrule 
\multirow{8}{*}{Citeseer} 
& GCN & 71.96±0.55 & 70.88±0.62 & 67.55±0.89 & 64.52±1.11 & 62.03±3.49\\ 
& GCN-SVD & 70.65±0.32 & 68.84±0.72 & 68.87±0.63 & 63.26±0.96 & 58.55±1.09\\ 
& GCN-Jaccard & 72.10±0.63 & 70.51±0.97 & 69.54±0.56 & 65.95±0.94 & 59.30±1.40\\ 
& SimP-GCN & \textbf{73.76.±0.78} & \textbf{73.12±0.85} & 72.38±0.67 & 71.75±1.54 & 69.37±1.50\\ 
& Pro-GNN & 73.28±0.69 & 72.93±0.57 & \textbf{72.51±0.75} & \textbf{72.03±1.11} & \textbf{70.02±2.28}\\ 
\cmidrule(l){2-7}
& GCN+F-MLP & \textbf{75.14±0.54} & \textbf{74.83±0.58} & 73.70±0.60 & 73.68±0.83 & \textbf{71.91±0.93}\\ 
& GCN+kNN-GCN & 74.80±0.58 & 74.76±0.56 & \textbf{73.79±0.48} & \textbf{73.86±0.79} & 71.74±1.34\\ 
\cmidrule(l){2-7}
& S-MLP & 69.31±0.34 & 68.90±0.40 & 62.41±0.65 & 62.11±0.38 & 54.99±0.31\\
& S-MLP+F-MLP & 74.31±0.47 & 74.12±0.38 & \textbf{74.82±0.22} & 74.60±0.58 & 69.30±0.69\\ 
& S-MLP+kNN-GCN & 
\textbf{74.94±0.41} & \textbf{75.04±0.23} & 74.53±0.24 & \textbf{75.01±0.22} & \textbf{72.80±0.74}\\ 
\midrule 
\midrule 
\multirow{8}{*}{Pubmed} 
& GCN & 87.19±0.09 & 83.09±0.13 & 81.21±0.09 & 78.66±0.12 & 77.35±0.19\\ 
& GCN-SVD & 83.44±0.21 & 83.41±0.15 & 83.27±0.21 & 83.10±0.18 & 83.01±0.22\\ 
& GCN-Jaccard & 87.06±0.06 & 86.39±0.06 & 85.70±0.07 & 84.76±0.08 & 83.88±0.05\\ 
& SimP-GCN & \textbf{87.59.±0.10} & 86.79±0.12 & 86.01±0.10 & 85.49±0.11 & 85.37±0.12\\ 
& Pro-GNN & 87.26±0.23 & \textbf{87.23±0.13} & \textbf{87.21±0.13} & \textbf{87.20±0.15} & \textbf{87.15±0.15}\\ 
\cmidrule(l){2-7}
& GCN+F-MLP & \textbf{87.62±0.05} & \textbf{86.98±0.09} & \textbf{86.85±0.09} & \textbf{86.05±0.08} & \textbf{86.04±0.04}\\ 
& GCN+kNN-CCN & 84.92±0.14 & 83.74±0.15 & 82.97±0.15 & 81.79±0.16 & 81.35±0.08\\
\cmidrule(l){2-7}
& S-MLP & 77.23±0.17 & 73.29±0.19 & 70.36±0.42 & 66.99±0.75 & 64.68±0.66\\
& S-MLP+F-MLP & \textbf{86.57±0.11} & \textbf{86.58±0.10} & \textbf{86.51±0.10} & \textbf{86.31±0.11} & \textbf{86.26±0.08}\\ 
& S-MLP+kNN-GCN & 81.90±0.06 & 81.51±0.08 & 81.41±0.09 & 80.88±0.11 & 80.28±0.08\\ 
\midrule 
\bottomrule 
\end{tabular}}%
\label{adversarial} 
\end{table}

\begin{figure}
\centering
\includegraphics[width=0.6\textwidth]{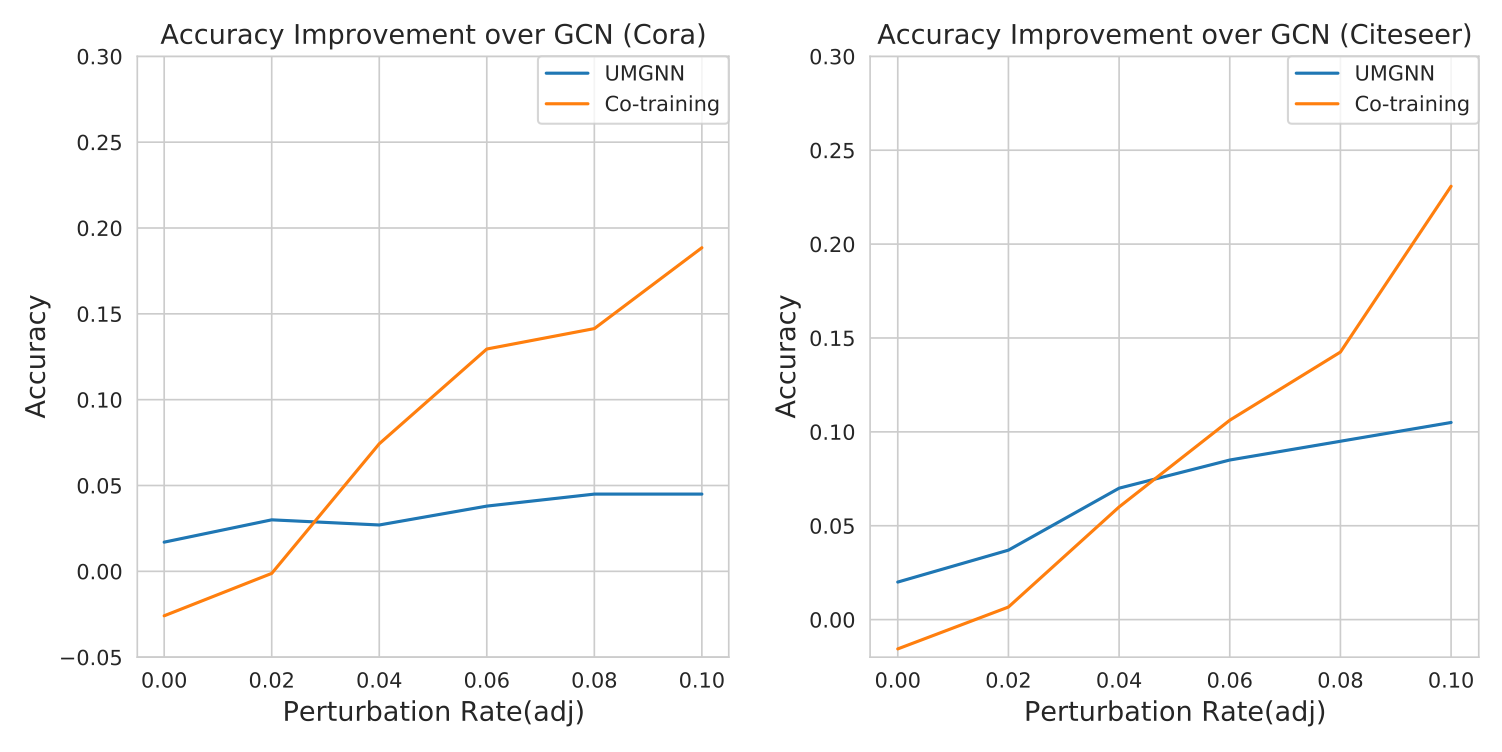}
\caption{Accuracy improvement over GCN. As perturbation rate increases, CoG~(GCN+MLP) outperforms UM-GNN by a large margin.} \label{fig3}
\vspace{-10pt}
\end{figure}

\subsection{Performance on clean graphs}
The performance of the ensemble models on clean graphs is shown in the 3rd column of Table~\ref{adversarial}. From the results, we make the following observations:
\begin{itemize}
    \item Our ensemble models outperform the baselines on clean graphs among three datasets. Specifically, the GCN+MLP model outperforms GCN by $0.59\%$, $3.18\%$, and $0.43\%$ for the three datasets respectively. For the other three defense models, the best results are $-0.52\%$, $1.28\%$, and $0.07\%$, respectively.
    \item Our ensemble models perform better than any single sub-model from the ensemble. For instance, the ensemble of GCN and kNN-GCN achieve an $84.27\%$ accuracy on Cora, while the accuracy is $83.50\%$ for GCN and $71.06\%$ for kNN-GCN.
\end{itemize}

Results show that our method achieves competitive performance on clean data, thus making it applicable in realistic settings where we have no idea if the graphs are perturbed.

\subsection{Performance under attacks}
Results under attack are shown in the 4th to 7th columns in Table~\ref{adversarial}. As we can see, the co-training framework effectively enhances the model's robustness against adversarial attacks. For example, the accuracy of GCN decreases drastically from $83.5\%$ to $59.6\%$ as the perturbation rate increases from 0 to $20\%$ on Cora. In comparison, the ensemble of GCN and kNN-GCN still achieves accuracy of $76.86\%$ even in the worst-perturbed case. Similar results are obtained on the other two datasets. 

Compared to the state-of-the-art defenses, our method improves the robustness among three datasets, especially on Cora and Citeseer. Pro-GNN~\cite{Jin2020} is the most robust model among the baselines. Therefore, we mainly compare our results with those of Pro-GNN. Our ensemble models outperform Pro-GNN by a large margin on Cora and Citeseer. Specifically, the combination of S-MLP and kNN-GCN gains $0.61\%$, $2.01\%$, $4.49\%$, and $7.43\%$ improvements as perturbation rate increases on Cora. The improvements on Citeseer are $1.86\%$, $2.65\%$, $2.02\%$ and $2.78\%$, respectively. On Pubmed, we also achieved comparable results. Moreover, it is worth noting that Pro-GNN trains slow~(over 100x longer than GCN), and requires lots of GPU memory~(more than 16GB for Pubmed). Our approach is much faster~(about 4x longer than GCN) and less memory-demanding~(same as GCN), making it more feasible in practical use. We also compare CoG with SimP-GCN~\cite{Jin2020_2}, which also focuses on exploiting the feature information. Results show that CoG achieves better performance among all three graphs and perturbation rates. Furthermore, the performance of SimP-GCN possesses high variance, especially on Cora. One possible reason is that SimP-GCN has unstable graph structure in its training stage. 

UM-GNN~\cite{shanthamallu2021uncertainty} uses different data splits from the above methods. It follows the original split in~\cite{Kipf2017} and uses perturbation rate from $0\%$ to $10\%$. Results in Fig.~\ref{fig3} show the accuracy improvement of CoG and UM-GNN over GCN. It shows that although UM-GNN performs slightly better than CoG when the perturbation rate is small, it is outperformed by a large margin as the perturbation rate increases. This is caused by the one-directional knowledge transferring scheme in UM-GNN. When the predictions of GNN is highly inaccurate, it will mislead the MLP model. This defect is avoided in CoG since CoG can transfer useful information in both direction. The performance of both GNN and MLP improves during the co-training process.

\subsection{Ablation Study}
In this section, we conduct an ablation study on the model calibration and label balancing techniques. We also evaluate how the number of iterations and the number of nodes to add in each iteration affect the performance.

\begin{figure*}
\centering
\includegraphics[width=\textwidth]{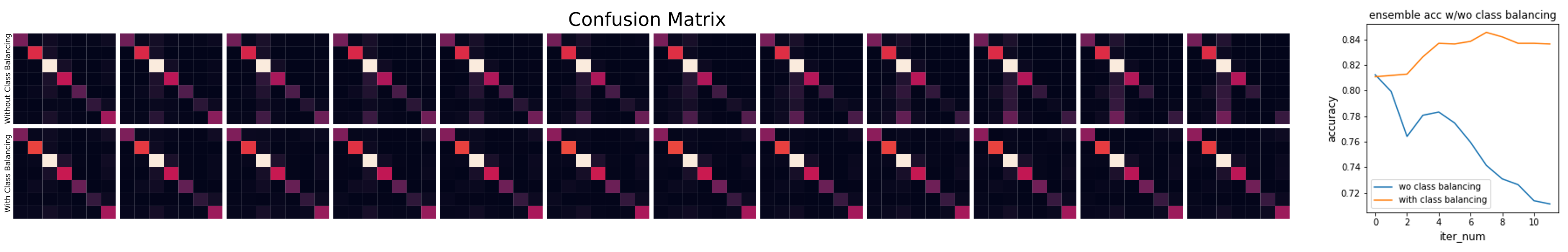}
\caption{Confusion matrix and accuracy of the ensemble model's prediction during the co-training process~(GCN+MLP with Cora). Without class balancing, the co-training process will overfit to the dominant class, thus leading to the decrease of accuracy as co-training continues.~(Results on Citeseer can be accessed in the supplementary material.)} \label{fig4}
\end{figure*}

\begin{figure}
\centering
\includegraphics[width=0.7\textwidth]{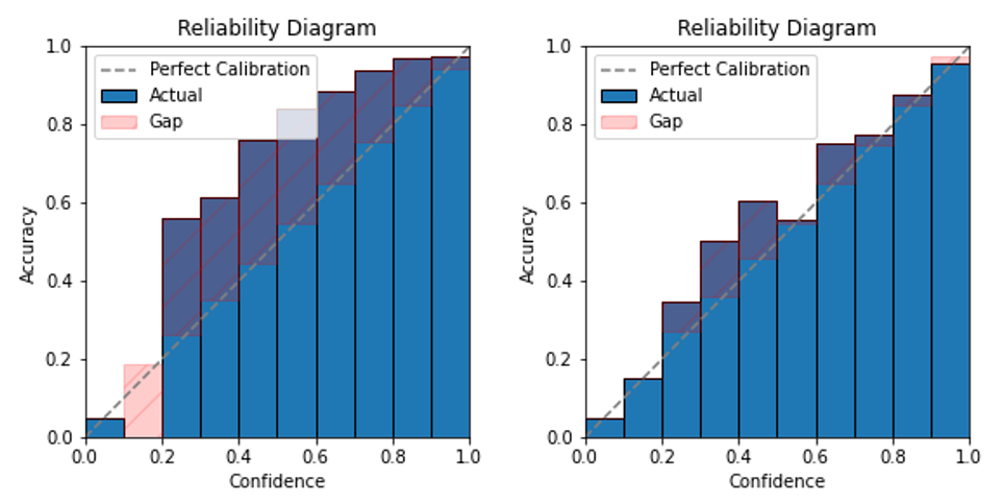}
\caption{Reliability Diagram of GCN w/o~(left) and with~(right)  calibration. After calibrating, model's output can better reflect its ground truth correctness likelihood.} \label{fig7}
\end{figure}

\begin{figure}
\centering
\includegraphics[width=0.8\textwidth]{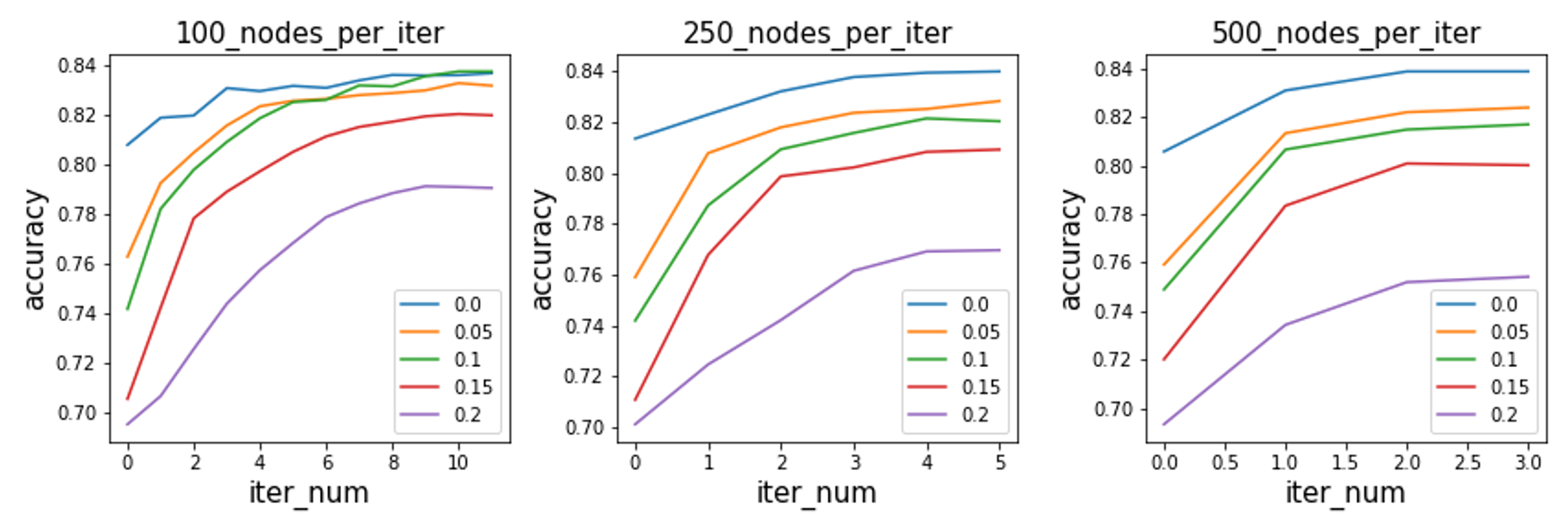}
\caption{Node classification accuracy using different hyper-parameters. Different colors denote the results under different perturbation rate.~(Results on Citeseer can be accessed in the supplementary material)} \label{fig5}
\end{figure}

\subsubsection{Model calibration and class balancing}
Fig.~\ref{fig7} shows the outputs of GCN with/without model calibration. The reliability diagram demonstrates that the vanilla GCN is under-confident. After temperature scaling, model's output can better reflect its the correctness likelihood of ground truth more accurately. To validate the benefit of the class balancing technique, we train an ensemble of GCN+MLP with/without class balancing on Cora and present the results in Fig.~\ref{fig4}. We report the confusion matrix and the accuracy of the ensemble model in each iteration. Results show that without class balancing, the co-training process will overfit to the dominant class. As the co-training process continues, more and more test data are labeled as the dominant class, thus impairing the performance of the ensemble model.

\subsubsection{Number of iterations and added nodes}
There are two hyper-parameters to set in CoG, which are the number of iterations and the number of nodes to add in each iteration. In our experiments, we add 100, 250, 500 nodes in each iteration respectively and evaluate CoG's performance form 0 iteration~(ensemble without co-training) to max
iterations~(add until no test data left). Results are shown in Fig.~\ref{fig5}. As the co-training process continues, the performance of CoG improves quickly in the beginning and stabilizes in the later iterations. Adding less nodes per iteration can slightly improve the performance of CoG. However, it also takes longer to train CoG, which means a trade-off between effectiveness and efficiency. Furthermore, comparing to ensemble sub-models without co-training~(the starting points), CoG can improve the performance by a large margin, especially on the perturbed data.

\begin{figure}
\centering
\includegraphics[width=0.6\textwidth]{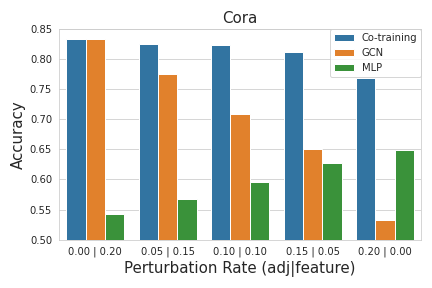}
\caption{Node classification accuracy of GCN, MLP and CoG under different ratios of feature and structure perturbations.~(Results on Citeseer can be accessed in the supplementary material)} \label{fig6}
\end{figure}

\subsection{Adaptive Attacks}
Evaluating CoG against adaptive attacks is necessary since attackers may attack CoG with both structure and feature perturbations in practice. However, adaptive attacks on CoG is non-trivial since the co-training process works in a non-differential manner. To evaluate CoG against adaptive attackers, we adapt the Metattack to attack the MLP model, keep the total perturbation budget as $20\%$, and evaluate CoG's performance under different ratios between feature and structure perturbations. Results are shown in Fig.~\ref{fig6}. We observe that CoG still achieves good robustness when both feature and structure information are perturbed.

\section{Conclusion}
In this paper, we present a co-training framework, named CoG, to learn an ensemble model that integrates the feature and the structure information of graph data. The orthogonal nature of these two views diversifies the outputs of sub-models and weakens the transferability of adversarial attacks between them.  Experiment results validate the effectiveness of CoG on both clean and perturbed graph. In addition, CoG can still achieve good robustness under the adaptive attack setting.

\newpage
\appendix

\section{Algorithm}
Here we demonstrate the overall algorithm of CoG in Algorithm~\ref{ag1}. The co-training process will stop once it reaches the preset number of iterations or there is no unlabeled data left. In the inference phase, we average each sub-model's outputs to obtain the predictions of their ensemble.

\begin{algorithm}[h]
\SetAlgoLined
 \KwInput{graph data $G=(X,A)$, nodes to add per iter $N^{add}$, Labeled dataset $S$, Unlabel dataset $U$, number of classes $C$}
 \KwOutput{$f_{struct}$, $f_{feat}$}
 Choose a structure-dominant model $f_{struct}$\;
 Choose a feature-dominant model $f_{feat}$\;
 \While{stopping condition not met}{
  Train $f_{struct}$ under structure view of $S$\;
  Train $f_{feat}$ under feature view of $S$\;
  Calculate confidence score $s_{struct}$ on $U$\;
  Calculate confidence score $s_{feat}$ on $U$\;
  \For{$c = 1$ to $C$}{
  Use $f_{struct}$ to label $N^{add}_{c}$ most confident nodes in $U$\;
  Use $f_{feat}$ to label $N^{add}_{c}$ most confident nodes in $U$\;
  Conducting temperature scaling for both sub-models\;
  \If{the same node is chosen}{
  Using the pseudo-label with higher confidence}
  }
  Update $S \& U$\;
 }
 \caption{Two-view co-training for graph data}\label{ag1}
\end{algorithm}

\section{Ensemble of structure-dominant models}
\begin{figure}[ht]
\centering
\includegraphics[width=.9\linewidth]{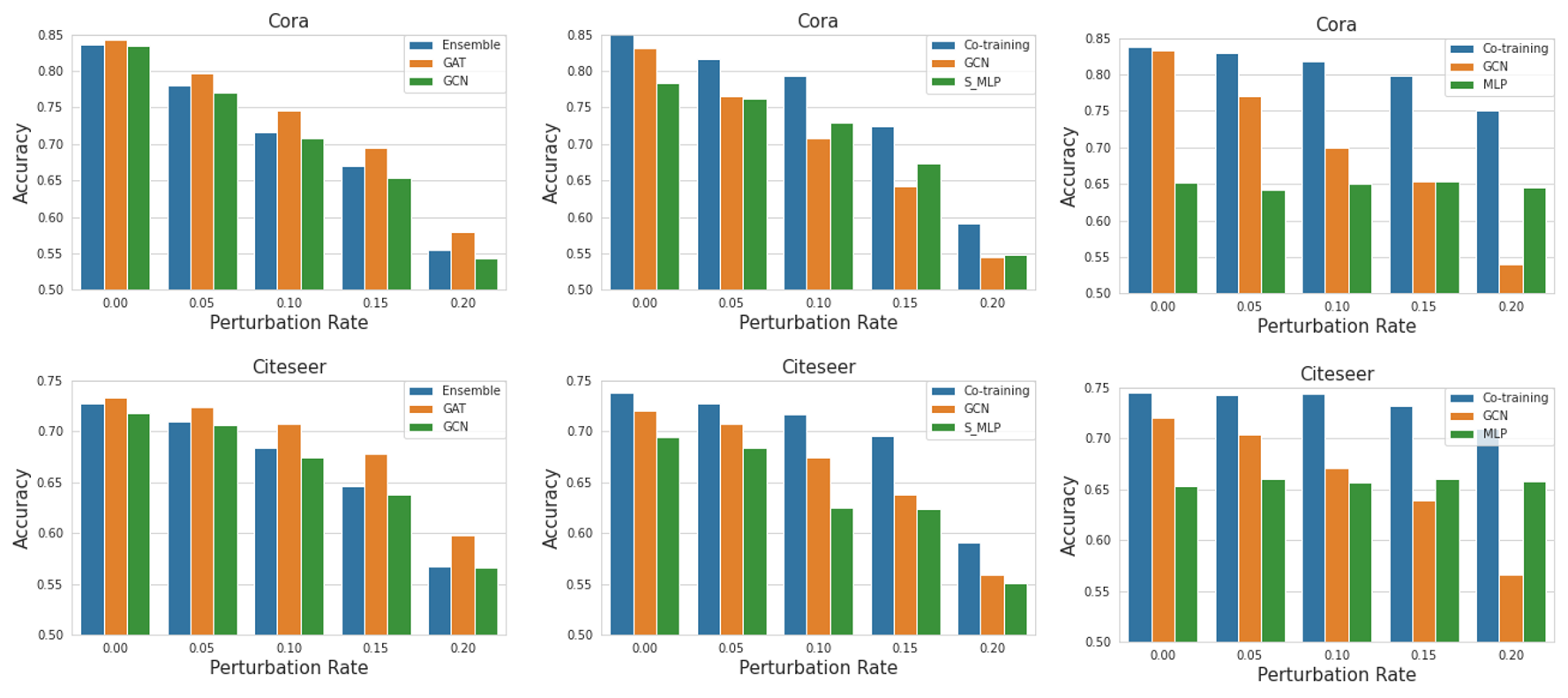}
\caption{Performance of ensemble of GCN+GAT, GCN+S\_MLP, GCN+MLP~(from left to right) under Metattack. Both the naive ensemble of GCN+GAT and the co-training of GCN and S\_MLP have little help for the robustness, compared to the co-traning of GCN+MLP.}
\label{fig:fig1}
\end{figure}

In the main paper, we show that a simple ensemble of GCN~\cite{Kipf2017}+GAT~\cite{Velickovic2018} still suffers from weak robustness against Metattack~\cite{zugner2019adversarial}. Fig.\ref{fig:fig1} shows the performance of GCN+GAT, GCN+S\_MLP and GCN+MLP on Cora and Citeseer. From the results we can conclude that the diversity of sub-models is crucial for the robustness of their ensemble.

\vspace{40pt}

\section{Robustness of CoG against PGD} \label{s3}
\begin{figure}[ht]
\centering
\includegraphics[width=.5\linewidth]{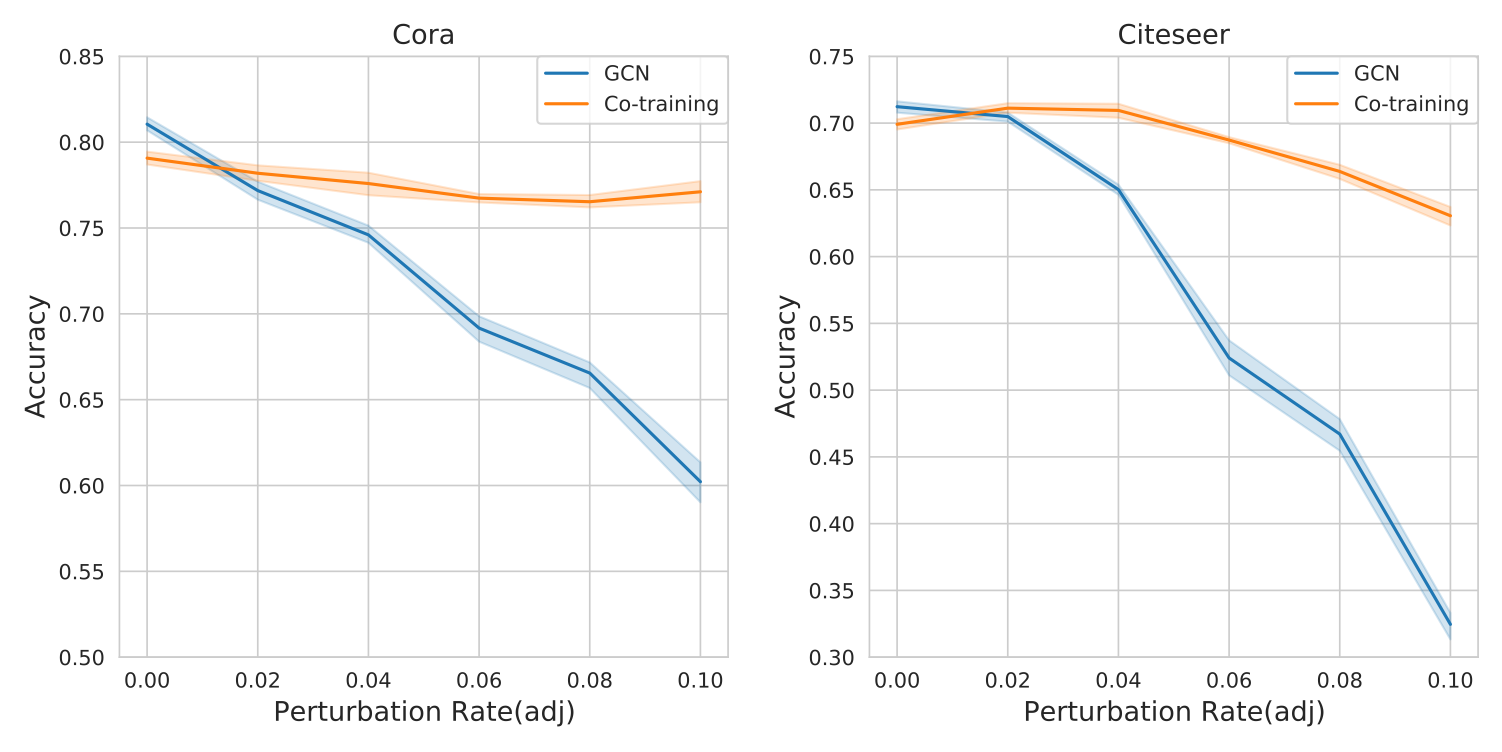}
\caption{The performance of CoG~(GCN+MLP) under PGD attack~(using the same data splits with UM-GNN).}
\label{fig:more}
\end{figure}

In the main paper, we mainly report CoG's performance against Metattack. Here we report its performance~(using GCN+MLP) against PGD~\cite{xu2019topology} in Fig.~\ref{fig:more}. As the results show, CoG also have good robustness under PGD attack.

\section{Comparing CoG and UM-GNN under PGD attack}
\begin{figure}[ht]
\centering
\includegraphics[width=.5\linewidth]{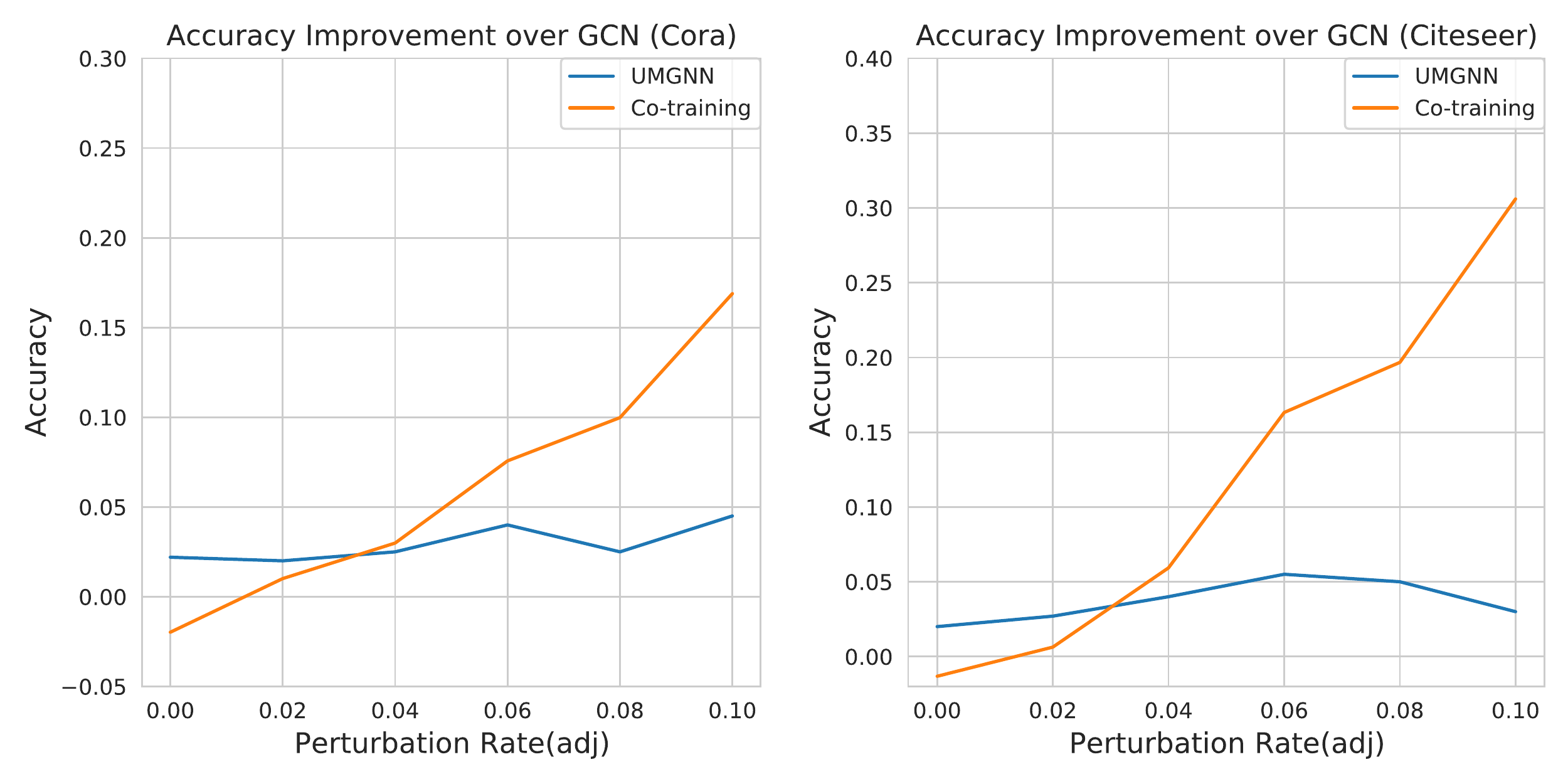}
\caption{Accuracy improvement over GCN under PGD attack. Results keep consistent with those under Metattack. As perturbation rate increases, CoG~(GCN+MLP) outperforms UM-GNN by a large margin.}
\label{fig:fig2}
\end{figure}

In main paper, we compare CoG and UM-GNN~\cite{shanthamallu2021uncertainty} under Metattack. Here we report the comparison under PGD attack in Fig.~\ref{fig:fig2}. Since the perturbed graph used in UM-GNN is not publicly available, to make a fair comparison, we report the performance improvement over GCN. For the comparison of absolute accuracy, please refer to the results in Section~\ref{s3} and \cite{shanthamallu2021uncertainty}.

\section{Model calibration}
\begin{figure}[ht]
\centering
\includegraphics[width=.6\linewidth]{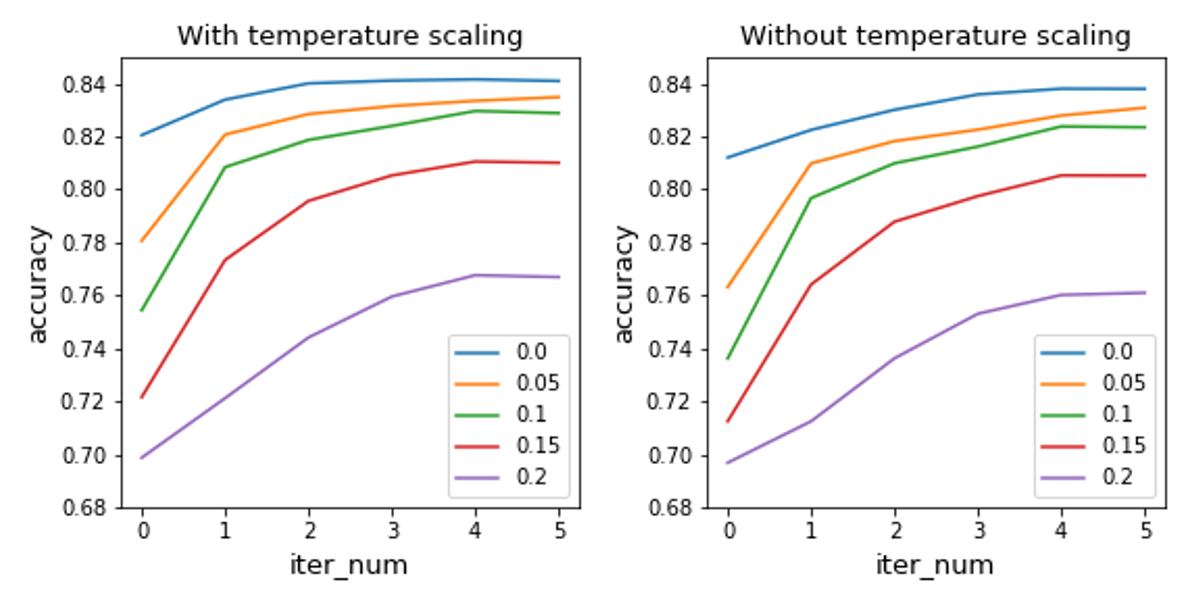}
\caption{Performance of CoG~(GCN+MLP on Cora) with/without calibration. With model calibration, CoG performs better, no matter the graph is perturbed or not.}
\label{fig:cali}
\end{figure}

In the main paper, we propose that model calibration is important for CoG. A calibrated classification model means that its softmax outputs~(confidence) should accurately measure the likelihood of correct predictions. In the main paper, we use the reliability diagram to demonstrate the miscalibration of GCN. Reliability diagram is a visual representation of model calibration~\cite{degroot1983comparison,niculescu2005predicting,guo2017calibration}. It splits the test data according to their output confidence, samples from them and plots the expected sample accuracy as a function of confidence. Here in Fig.~\ref{fig:cali}, we show the performance improvement with model calibration. Using temperature scaling to calibrate the model brings a $1\sim2\%$ performance improvement. 

\section{Class balancing}
\begin{figure}[ht]
\centering
\includegraphics[width=.8\linewidth]{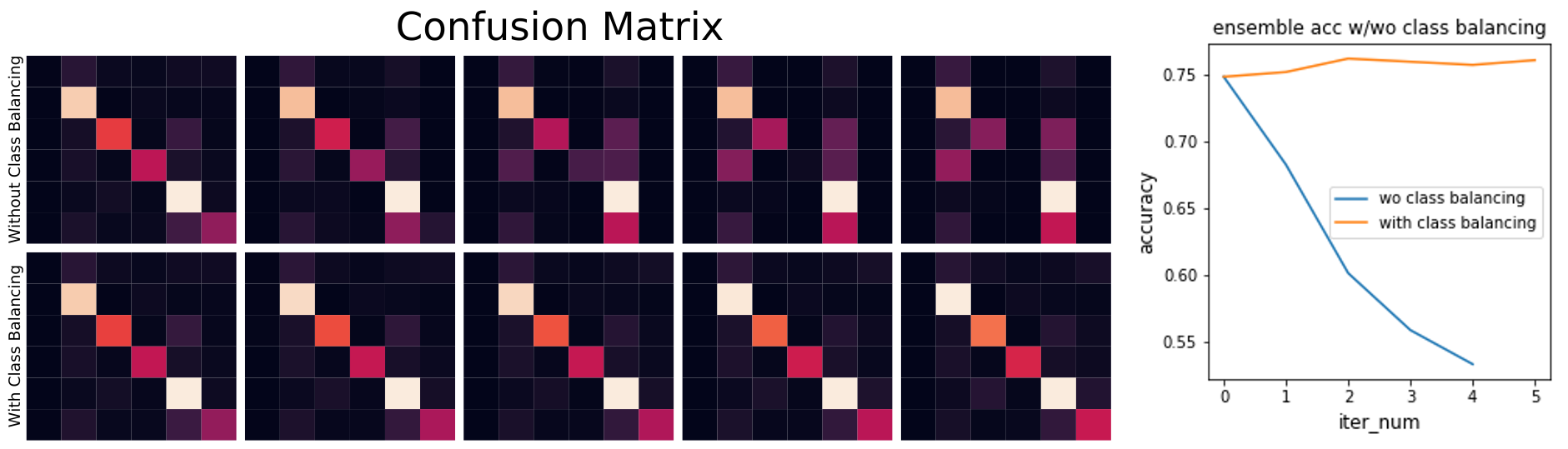}
\caption{Confusion matrix and accuracy of the ensemble model's prediction during the co-training process~(GCN+MLP with unperturbed Citeseer data).}
\label{fig:class}
\end{figure}

\begin{figure}[ht]
\centering
\includegraphics[width=.8\linewidth]{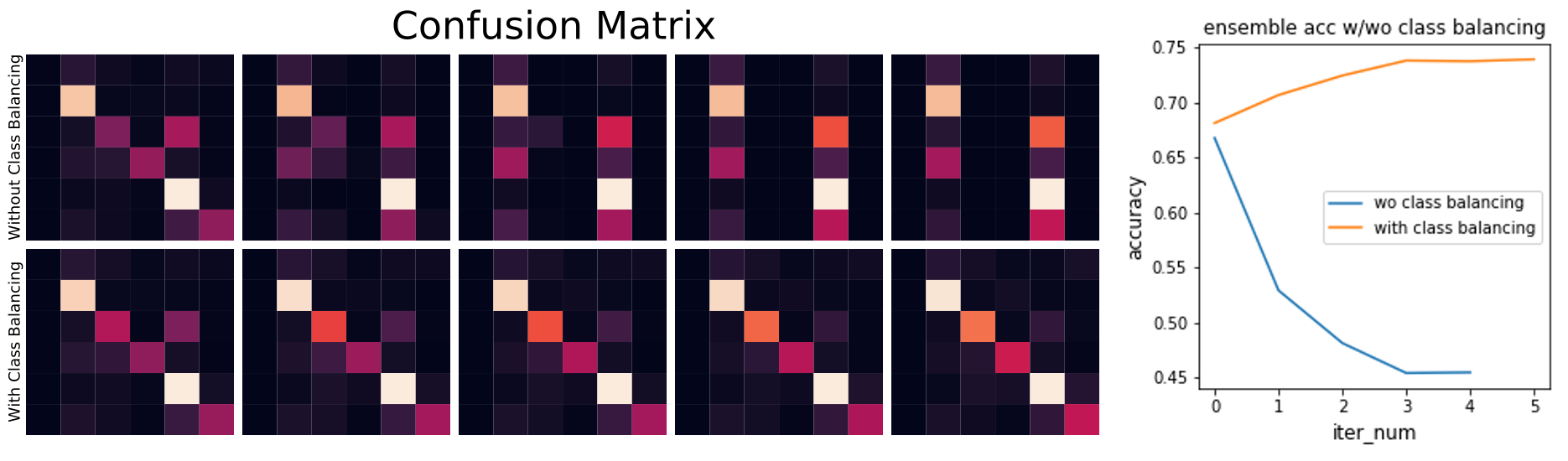}
\caption{Confusion matrix and accuracy of the ensemble model's prediction during the co-training process~(GCN+MLP with 20\% perturbed Citeseer data).}
\label{fig:class2}
\end{figure}

In the main paper, we compare CoG's performance with/without classing balancing on Cora. Results show that without class balancing, the ensemble model will overfit to the dominant class. Here we report the confusion matrix and ensemble accuracy on Citeseer in Fig.~\ref{fig:class} and Fig.~\ref{fig:class2}. Confusion matrix is a specific layout for visualizing the performance of model's predictions. Each row of the matrix represents the instances in an actual class while each column represents the instances in a predicted class. Instances in the diagonal of the matrix are correctly classified. For more details about the confusion matrix, we refer the reviewers to \cite{powers2020evaluation}. As we can see in Fig.~\ref{fig:class}, class 5 is the dominant class in this dataset. Without class balancing, more and more instances from other classes are misclassified as class 5 as the co-training process continues, leading to an overfitting to class 5 and a decrease in accuracy. We obtain the same results on the perturbed data. From the confusion matrix in Fig.~\ref{fig:class2}, we can also see how the co-training process~(with class balancing) corrects the misclassified instances.

\section{Hyper-parameters}
\begin{figure}[ht]
\centering
\includegraphics[width=.7\linewidth]{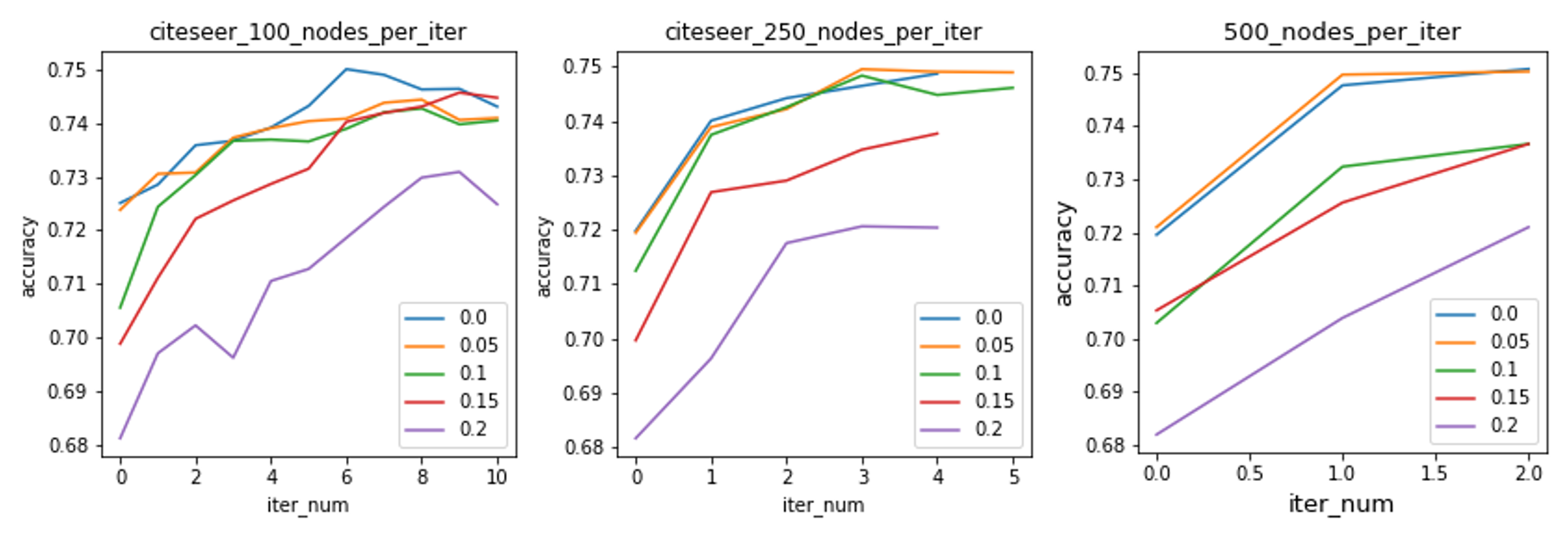}
\caption{Node classification accuracy using different hyper-parameters. Generally, adding less nodes per iteration leads to better accuracy, although it is more time-demanding.}
\label{fig:hyper}
\end{figure}

In main paper, we compare CoG's performance with different hyper-parameters~(nodes to add per iteration \& number of iterations) on Cora. Here we report the results on Citeseer in Fig.~\ref{fig:hyper}. Results are consistent with those on Cora: with less nodes to add per iterations, the final accuracy is slightly higher. Note that it is also more time-demanding. As for co-training iterations, performance improves quickly in the beginning and stabilizes in the final stage.

\section{Adaptive attack}
\begin{figure}[ht]
\centering
\includegraphics[width=.5\linewidth]{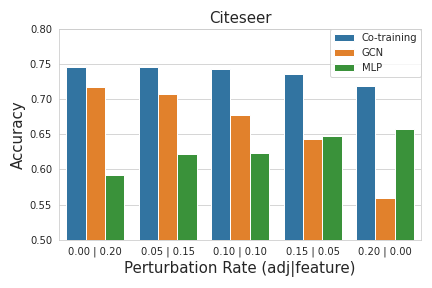}
\caption{Node classification accuracy of GCN, MLP and CoG under different ratios of feature and structure perturbations. The total perturbation budget is 20\%.}
\label{fig:adaptive}
\end{figure}

In main paper, we demonstrate CoG's performance under adaptive attacks on Cora. Here we report the results on Citeseer in Fig.~\ref{fig:adaptive}. CoG can still achieve good robustness when the node features and graph structures are perturbed simultaneously. 

\bibliographystyle{splncs04}
\bibliography{ref}

\begin{thebibliography}{10}
\providecommand{\url}[1]{\texttt{#1}}
\providecommand{\urlprefix}{URL }
\providecommand{\doi}[1]{https://doi.org/#1}

\bibitem{blum1998combining}
Blum, A., Mitchell, T.: Combining labeled and unlabeled data with co-training.
  In: Proceedings of the eleventh annual conference on Computational learning
  theory. pp. 92--100 (1998)

\bibitem{Bojchevski2019}
Bojchevski, A., G{\"u}nnemann, S.: Adversarial attacks on node embeddings via
  graph poisoning. In: International Conference on Machine Learning. pp.
  695--704. PMLR (2019)

\bibitem{Breiman1996}
Breiman, L.: Bagging predictors. Machine learning  \textbf{24}(2),  123--140
  (1996)

\bibitem{chen2021enhancing}
Chen, L., Li, X., Wu, D.: Enhancing robustness of graph convolutional networks
  via dropping graph connections. In: Machine Learning and Knowledge Discovery
  in Databases: European Conference, ECML PKDD 2020, Ghent, Belgium, September
  14--18, 2020, Proceedings, Part III. pp. 412--428. Springer International
  Publishing (2021)

\bibitem{chung1997spectral}
Chung, F.R., Graham, F.C.: Spectral graph theory. No.~92, American Mathematical
  Soc. (1997)

\bibitem{dai2018adversarial}
Dai, H., Li, H., Tian, T., Huang, X., Wang, L., Zhu, J., Song, L.: Adversarial
  attack on graph structured data. In: International conference on machine
  learning. pp. 1115--1124. PMLR (2018)

\bibitem{degroot1983comparison}
DeGroot, M.H., Fienberg, S.E.: The comparison and evaluation of forecasters.
  Journal of the Royal Statistical Society: Series D (The Statistician)
  \textbf{32}(1-2),  12--22 (1983)

\bibitem{dietterich2000ensemble}
Dietterich, T.G.: Ensemble methods in machine learning. In: International
  workshop on multiple classifier systems. pp. 1--15. Springer (2000)

\bibitem{Entezari2020}
Entezari, N., Al{-}Sayouri, S.A., Darvishzadeh, A., Papalexakis, E.E.: All you
  need is low (rank): Defending against adversarial attacks on graphs. In:
  Caverlee, J., Hu, X.B., Lalmas, M., Wang, W. (eds.) {WSDM} '20: The
  Thirteenth {ACM} International Conference on Web Search and Data Mining,
  Houston, TX, USA, February 3-7, 2020. pp. 169--177. {ACM} (2020).
  \doi{10.1145/3336191.3371789}, \url{https://doi.org/10.1145/3336191.3371789}

\bibitem{Fey2019}
Fey, M., Lenssen, J.E.: {Fast Graph Representation Learning with PyTorch
  Geometric} (1), ~1--9 (2019), \url{http://arxiv.org/abs/1903.02428}

\bibitem{guo2017calibration}
Guo, C., Pleiss, G., Sun, Y., Weinberger, K.Q.: On calibration of modern neural
  networks. In: International Conference on Machine Learning. pp. 1321--1330.
  PMLR (2017)

\bibitem{Hamilton2017}
Hamilton, W.L., Ying, Z., Leskovec, J.: Inductive representation learning on
  large graphs. In: Guyon, I., von Luxburg, U., Bengio, S., Wallach, H.M.,
  Fergus, R., Vishwanathan, S.V.N., Garnett, R. (eds.) Advances in Neural
  Information Processing Systems 30: Annual Conference on Neural Information
  Processing Systems 2017, December 4-9, 2017, Long Beach, CA, {USA}. pp.
  1024--1034 (2017),
  \url{https://proceedings.neurips.cc/paper/2017/hash/5dd9db5e033da9c6fb5ba83c7a7ebea9-Abstract.html}

\bibitem{Hansen1990}
Hansen, L.K., Salamon, P.: {Neural Network Ensembles}. IEEE Transactions on
  Pattern Analysis and Machine Intelligence  (1990). \doi{10.1109/34.58871}

\bibitem{Jin2020_2}
Jin, W., Derr, T., Wang, Y., Ma, Y., Liu, Z., Tang, J.: {Node Similarity
  Preserving Graph Convolutional Networks}  (2020),
  \url{http://arxiv.org/abs/2011.09643}

\bibitem{Jin2020}
Jin, W., Ma, Y., Liu, X., Tang, X., Wang, S., Tang, J.: Graph structure
  learning for robust graph neural networks. In: Gupta, R., Liu, Y., Tang, J.,
  Prakash, B.A. (eds.) {KDD} '20: The 26th {ACM} {SIGKDD} Conference on
  Knowledge Discovery and Data Mining, Virtual Event, CA, USA, August 23-27,
  2020. pp. 66--74. {ACM} (2020),
  \url{https://dl.acm.org/doi/10.1145/3394486.3403049}

\bibitem{Kariyappa2019}
Kariyappa, S., Qureshi, M.K.: {Improving adversarial robustness of ensembles
  with diversity training}. arXiv  (2019)

\bibitem{Kipf2017}
Kipf, T.N., Welling, M.: Semi-supervised classification with graph
  convolutional networks. In: 5th International Conference on Learning
  Representations, {ICLR} 2017, Toulon, France, April 24-26, 2017, Conference
  Track Proceedings. OpenReview.net (2017),
  \url{https://openreview.net/forum?id=SJU4ayYgl}

\bibitem{Kuncheva2003}
Kuncheva, L.I., Whitaker, C.J.: {Measures of diversity in classifier ensembles
  and their relationship with the ensemble accuracy}. Machine Learning  (2003).
  \doi{10.1023/A:1022859003006}

\bibitem{li2020deeprobust}
Li, Y., Jin, W., Xu, H., Tang, J.: Deeprobust: A pytorch library for
  adversarial attacks and defenses (2020)

\bibitem{Luo2020}
Luo, D., Cheng, W., Yu, W., Zong, B., Ni, J., Chen, H., Zhang, X.: {Learning to
  drop: Robust graph neural network via topological denoising}. arXiv  (2020)

\bibitem{mccallum2000automating}
McCallum, A.K., Nigam, K., Rennie, J., Seymore, K.: Automating the construction
  of internet portals with machine learning. Information Retrieval
  \textbf{3}(2),  127--163 (2000)

\bibitem{niculescu2005predicting}
Niculescu-Mizil, A., Caruana, R.: Predicting good probabilities with supervised
  learning. In: Proceedings of the 22nd international conference on Machine
  learning. pp. 625--632 (2005)

\bibitem{Pang2019}
Pang, T., Xu, K., Du, C., Chen, N., Zhu, J.: Improving adversarial robustness
  via promoting ensemble diversity. In: Chaudhuri, K., Salakhutdinov, R. (eds.)
  Proceedings of the 36th International Conference on Machine Learning, {ICML}
  2019, 9-15 June 2019, Long Beach, California, {USA}. Proceedings of Machine
  Learning Research, vol.~97, pp. 4970--4979. {PMLR} (2019),
  \url{http://proceedings.mlr.press/v97/pang19a.html}

\bibitem{powers2020evaluation}
Powers, D.M.: Evaluation: from precision, recall and f-measure to roc,
  informedness, markedness and correlation. arXiv preprint arXiv:2010.16061
  (2020)

\bibitem{sen2008collective}
Sen, P., Namata, G., Bilgic, M., Getoor, L., Galligher, B., Eliassi-Rad, T.:
  Collective classification in network data. AI magazine  \textbf{29}(3),
  93--93 (2008)

\bibitem{shanthamallu2021uncertainty}
Shanthamallu, U.S., Thiagarajan, J.J., Spanias, A.: Uncertainty-matching graph
  neural networks to defend against poisoning attacks. In: Proceedings of the
  AAAI Conference on Artificial Intelligence. vol.~35, pp. 9524--9532 (2021)

\bibitem{ensemble}
Stellargraph: Ensemble models for node classification (2020),
  \url{https://stellargraph.readthedocs.io/en/stable/demos/ensembles/ensemble-node-classification-example.html}

\bibitem{teixeira2019graph}
Teixeira, L., Jalaian, B., Ribeiro, B.: Are graph neural networks
  miscalibrated? arXiv preprint arXiv:1905.02296  (2019)

\bibitem{tramer2017space}
Tram{\`e}r, F., Papernot, N., Goodfellow, I., Boneh, D., McDaniel, P.: The
  space of transferable adversarial examples. arXiv preprint arXiv:1704.03453
  (2017)

\bibitem{Velickovic2018}
Velickovic, P., Cucurull, G., Casanova, A., Romero, A., Li{\`{o}}, P., Bengio,
  Y.: Graph attention networks. In: 6th International Conference on Learning
  Representations, {ICLR} 2018, Vancouver, BC, Canada, April 30 - May 3, 2018,
  Conference Track Proceedings. OpenReview.net (2018),
  \url{https://openreview.net/forum?id=rJXMpikCZ}

\bibitem{von2007tutorial}
Von~Luxburg, U.: A tutorial on spectral clustering. Statistics and computing
  \textbf{17}(4),  395--416 (2007)

\bibitem{wang2007analyzing}
Wang, W., Zhou, Z.H.: Analyzing co-training style algorithms. In: European
  conference on machine learning. pp. 454--465. Springer (2007)

\bibitem{Wu2019}
Wu, H., Wang, C., Tyshetskiy, Y., Docherty, A., Lu, K., Zhu, L.: Adversarial
  examples for graph data: Deep insights into attack and defense. In: Kraus, S.
  (ed.) Proceedings of the Twenty-Eighth International Joint Conference on
  Artificial Intelligence, {IJCAI} 2019, Macao, China, August 10-16, 2019. pp.
  4816--4823. ijcai.org (2019). \doi{10.24963/ijcai.2019/669},
  \url{https://doi.org/10.24963/ijcai.2019/669}

\bibitem{xu2019topology}
Xu, K., Chen, H., Liu, S., Chen, P.Y., Weng, T.W., Hong, M., Lin, X.: Topology
  attack and defense for graph neural networks: An optimization perspective.
  arXiv preprint arXiv:1906.04214  (2019)

\bibitem{Yang2020}
Yang, H., Zhang, J., Dong, H., Inkawhich, N., Gardner, A., Touchet, A., Wilkes,
  W., Berry, H., Li, H.: {DVERGE: Diversifying vulnerabilities for enhanced
  robust generation of ensembles}. arXiv (NeurIPS),  1--20 (2020)

\bibitem{Zhang2020}
Zhang, X., Zitnik, M.: {GNNGuard: Defending Graph Neural Networks against
  Adversarial Attacks}  (2020), \url{http://arxiv.org/abs/2006.08149}

\bibitem{zugner2018adversarial}
Z{\"u}gner, D., Akbarnejad, A., G{\"u}nnemann, S.: Adversarial attacks on
  neural networks for graph data. In: Proceedings of the 24th ACM SIGKDD
  International Conference on Knowledge Discovery \& Data Mining. pp.
  2847--2856 (2018)

\bibitem{zugner2019adversarial}
Z{\"{u}}gner, D., G{\"{u}}nnemann, S.: Adversarial attacks on graph neural
  networks via meta learning. In: 7th International Conference on Learning
  Representations, {ICLR} 2019, New Orleans, LA, USA, May 6-9, 2019.
  OpenReview.net (2019), \url{https://openreview.net/forum?id=Bylnx209YX}

\end{thebibliography}
\end{document}